\documentclass[sigconf]{acmart}
\usepackage{balance}
\hypersetup{draft}

\AtBeginDocument{%
  }

\setcopyright{acmlicensed}
\copyrightyear{2025}
\acmYear{2025}
\acmDOI{10.1145/YYYYYYY.YYYYYYY}

\acmConference[ACMSE 2025]{2025 ACM Southeast Conference}{April 24--26, 2025}{Cape Girardeau, MO, USA}
\acmBooktitle{2025 ACM Southeast Conference (ACMSE 2025), April 24--26, 2025, Cape Girardeau, MO, USA}
\acmPrice{15.00}
\acmISBN{979-8-4007-1277-7/25/04}

\begin{document}
\pagestyle{empty}

\title{Meta-Reinforcement Learning with Discrete World Models for Adaptive Load Balancing}

\author{Cameron Redovian}
\affiliation{
  \institution{Kennesaw State University}
  \city{Marietta}
  \state{Georgia}
  \country{USA}
}
\email{credovia@students.kennesaw.edu}

\renewcommand{\shortauthors}{Redovian}

\begin{abstract}
We integrate a meta-reinforcement learning algorithm with the DreamerV3 architecture to improve load balancing in operating systems. This approach enables rapid adaptation to dynamic workloads with minimal retraining, outperforming the Advantage Actor-Critic (A2C) algorithm in standard and adaptive trials. It demonstrates robust resilience to catastrophic forgetting, maintaining high performance under varying workload distributions and sizes. These findings have important implications for optimizing resource management and performance in modern operating systems. By addressing the challenges posed by dynamic and heterogeneous workloads, our approach advances the adaptability and efficiency of reinforcement learning in real-world system management tasks.
\end{abstract}



\maketitle

\section{Research Statement and Conjecture}
Building on the foundational work by Meyer et al. \cite{meyer2023harnessing}, which demonstrated the potential of discrete representations to mitigate catastrophic forgetting, this study extends the DreamerV3 architecture by incorporating a recurrent policy network. This enhancement enables the model to rapidly adapt to novel load distributions, addressing a key challenge in reinforcement learning for dynamic environments. By leveraging discrete world representations, our approach minimizes the need for extensive retraining, offering a scalable and efficient solution for meta-reinforcement learning (meta-RL). Using the Park operating system environments \cite{mao2019park} for realistic benchmarking, this study evaluates the practicality of these methods in adaptive load balancing tasks. We hypothesize that integrating the RL² algorithm with the DreamerV3 model will significantly outperform traditional reinforcement learning algorithms. Specifically, we expect the proposed approach to exhibit rapid convergence to optimal solutions, robust adaptation to workload variability, and a marked reduction in catastrophic forgetting, even under dynamic and unpredictable conditions.

\section{Related Work}

\subsection{Meta-RL with Recurrent Policy Networks}
Meta-reinforcement learning (meta-RL) enables agents to rapidly adapt to new tasks by leveraging prior experience. One prominent approach utilizes recurrent policy networks, which encode past interactions within their hidden states, effectively allowing the agent to "remember" past experiences and adjust its behavior accordingly.

Early foundational work by Duan et al. \cite{duan2016rl} introduced RL², a framework where reinforcement learning itself is embedded within a recurrent neural network (RNN). This approach allows the agent to internalize task-specific information through recurrent state updates, leading to rapid adaptation without requiring explicit model updates. Wang et al. \cite{wang2016learning} independently explored a similar paradigm, demonstrating that an RNN-based policy could act as an implicit reinforcement learning algorithm, adjusting its strategy within an episode based on past observations and rewards.

These works established the viability of memory-based adaptation, where recurrent architectures enable policies to generalize across task distributions without requiring explicit retraining. The success of these methods underscored the importance of structured memory in meta-RL, particularly in dynamic environments where task distributions evolve over time.

\subsection{Gradient-Based Meta-RL}
While recurrent architectures provide implicit adaptation, gradient-based meta-learning methods offer an alternative approach by optimizing initial policy parameters for rapid fine-tuning. Finn et al. \cite{finn2017model} introduced Model-Agnostic Meta-Learning (MAML), which explicitly trains policy parameters to be highly adaptable with minimal gradient updates. This approach differs from recurrent meta-learning in that it does not rely on hidden states for adaptation but instead ensures that the policy can quickly adjust to new tasks through learned initialization.

Although MAML has proven effective in few-shot learning, its reliance on explicit gradient updates poses challenges in non-stationary environments, where continual adaptation is required without retraining. Recent work has sought to bridge the gap between recurrent and gradient-based methods, combining memory-driven task inference with rapid parameter updates to enhance meta-RL performance in dynamic settings.

\subsection{Meta-RL for Continual Learning}
Meta-RL is particularly well-suited for continual learning, where agents must handle sequences of tasks while mitigating catastrophic forgetting. Nagabandi et al. \cite{nagabandi2018learning} demonstrated the effectiveness of meta-learning in online adaptation, enabling robots to recover from unexpected perturbations such as mechanical failures or environmental changes. Their approach meta-trained a model-based RL agent to adapt its dynamics model online, facilitating fast recovery from non-stationary conditions.

Zintgraf et al. \cite{zintgraf2019varibad} extended this idea with VariBAD, a Bayes-adaptive meta-RL framework that uses variational inference to maintain a belief distribution over tasks. By leveraging a latent task embedding, VariBAD enables agents to balance exploration and exploitation in continually changing environments, improving adaptation efficiency.

A more recent advancement, CoMPS (Continual Meta Policy Search) by Berseth et al. \cite{berseth2021comps}, addresses the challenge of long-term meta-learning, where tasks arrive sequentially without revisitation. Unlike traditional meta-RL methods that assume access to all training tasks upfront, CoMPS continually updates its meta-policy across task sequences, reducing catastrophic forgetting while preserving adaptability.

While recurrent meta-learning enables quick task inference, recent research has examined its limitations in a continual setting. One perspective reformulates recurrent meta-RL as a partially observable Markov decision process (MDP), suggesting that the RNN’s hidden state serves as a belief state over tasks \cite{alver2021goinginsiderecurrentmeta}. This insight aligns with approaches like VariBAD and explains how recurrent policies can implement a form of task inference. However, other work has shown that recurrent policies alone are not sufficient to prevent forgetting in long task sequences. For example, a 2024 study by de Lara et al. found that even though recurrence “improves adaptability, it still suffers from catastrophic forgetting” when tasks change sequentially. The agent’s performance on past tasks can deteriorate as it focuses on the current one, indicating that additional mechanisms (such as regularization, memory replay, or meta-learned weight updates) are necessary to truly achieve continual learning \cite{de2024recurrent}. 

\subsection{Discrete World Models}
The DreamerV3 architecture \cite{hafner2023mastering} has emerged as a transformative advancement in reinforcement learning through its integration of world models with discrete latent spaces. Unlike conventional reinforcement learning approaches that directly operate on raw observations, DreamerV3 models the environment as a compact, interpretable latent space, enabling agents to anticipate future states and make decisions based on these predictions. These latent spaces distill the high-dimensional complexity of environmental dynamics into actionable representations, drastically improving sample efficiency and generalization across tasks.

Discrete world models have demonstrated immense potential in continual learning, as highlighted by Meyer et al. \cite{meyer2023harnessing}. Their research underscores the stability and efficiency that discrete representations bring to reinforcement learning agents operating in dynamic environments. By improving task differentiation, these models enable agents to better delineate between old and new tasks, reducing catastrophic forgetting—a persistent challenge in lifelong learning scenarios.

In operating system tasks like load balancing, where workloads are unpredictable and subject to frequent shifts, discrete latent spaces offer a critical advantage. They allow agents to retain robust state representations despite external variability, ensuring that prior knowledge is preserved while adapting to new conditions. This capability not only reduces the need for extensive retraining but also facilitates faster adaptation, making it a practical solution for high-stakes, real-time decision-making environments. By leveraging the stability and generalization properties of discrete world models, this study seeks to address the dual challenges of catastrophic forgetting and adaptability, establishing a more resilient approach to dynamic workload management.

\subsection{Park OS Environments}
The Park platform \cite{mao2019park} is a benchmark suite for reinforcement learning research in operating system (OS) domains, providing standardized environments for tasks such as load balancing, scheduling, and resource allocation. For environments with well-understood dynamics, such as the load balancing environment used in this research, Park provides high-fidelity simulators that closely mimic real-world systems by leveraging historical data and real-world workload traces. This approach ensures that the simulated environments reflect realistic task distributions, load fluctuations, and scheduling constraints encountered in production systems. By doing so, Park mitigates the limitations of synthetic benchmarks, providing a realistic yet controlled evaluation platform for reinforcement learning agents. 

In the context of this study, Park provides an essential framework for evaluating the adaptability and generalization capabilities of the DreamerV3-based RL² agent. However, using Park as a benchmark is not without challenges. Many of its resources, such as preconfigured task files, are outdated or
inaccessible, necessitating significant modifications to the environment. Despite these limitations, Park remains an invaluable tool for testing the scalability of reinforcement learning methods in
complex, dynamic systems. By integrating dynamic workload distributions into Park’s environments, this project ensures a rigorous assessment of the proposed methods, highlighting their potential
for real-world applicability.

\section{Methodology}

\subsection{Implementation Overview}
Our study builds upon the DreamerV3 architecture \cite{hafner2023mastering}, leveraging its world model-based reinforcement learning approach for adaptive load balancing in operating systems. We implemented DreamerV3 in PyTorch, ensuring full compatibility with the Park OS simulation environment \cite{mao2019park}. To establish a baseline, we compared the performance of our agent against Advantage Actor-Critic (A2C) \cite{pmlr-v48-mniha16}, which was originally used in the Park benchmark.

\subsection{DreamerV3 with Recurrent Policy}
While DreamerV3 achieves state-of-the-art performance across various domains, its policy network lacks explicit memory mechanisms to handle dynamic environments where task distributions change over time. To enhance meta-reinforcement learning capabilities, we augmented the DreamerV3 policy with a Gated Recurrent Unit (GRU) layer, drawing inspiration from RL² \cite{duan2016rl}. This modification allows the agent to leverage temporal dependencies in action selection, improving adaptation to shifting workload conditions without requiring additional retraining.

The modified DreamerV3 architecture consists of the following components:

\begin{itemize}
    \item \textbf{World Model:} Encodes the environment into a discrete latent space, predicting future states using a recurrent sequence model.
    \item \textbf{Policy Network:} Receives latent state representations from the world model and produces actions.
    \item \textbf{GRU Augmentation:} A 256-unit GRU layer is inserted before the final action selection step, allowing the agent to retain short-term memory of workload fluctuations.
    \item \textbf{Actor and Critic Heads:} Two separate MLPs receive GRU outputs to produce policy logits (actor) and value estimates (critic).
\end{itemize}

Formally, let $z_t$ denote the latent state representation at timestep $t$, produced by the world model. The modified policy function is defined as:

\begin{equation}
    \begin{split}
        h_t &= \text{GRU}(h_{t-1}, z_t)  \\
        a_t &\sim \pi_\theta(h_t)
    \end{split}
\end{equation}

\noindent where $h_t$ is the hidden state of the GRU, $\pi_\theta$ is the policy function parameterized by a neural network, and $a_t$ is the action selected according to the policy at timestep $t$.

The GRU enables recurrent information flow, allowing the policy to adjust based on recent workload patterns without relying solely on the world model’s predictions. This is particularly beneficial in non-stationary environments, where changes in job arrival rates and job sizes require rapid adaptation.

DreamerV3’s hyperparameters remained unchanged from the original implementation \cite{hafner2023mastering}, except for the addition of the GRU layer to the policy network. Table~\ref{tab:hyperparams} summarizes the key hyperparameters.

\begin{table}[ht]
    \centering
    \caption{DreamerV3 Hyperparameters}
    \label{tab:hyperparams}
    \begin{tabular}{lc}
        \toprule
        \textbf{Parameter} & \textbf{Value} \\
        \midrule
        Replay Capacity & $5 \times 10^6$ \\
        Batch Size & 16 \\
        Batch Length & 64 \\
        Learning Rate & $4 \times 10^{-5}$ \\
        Gradient Clipping & AGC(0.3) \\
        Optimizer & LaProp ($\epsilon=10^{-20}$) \\
        \midrule
        \textbf{World Model} & \\
        Reconstruction Loss Scale ($\beta_{pred}$) & 1 \\
        Dynamics Loss Scale ($\beta_{dyn}$) & 1 \\
        Representation Loss Scale ($\beta_{rep}$) & 0.1 \\
        Latent Unimix & 1\% \\
        Free Nats & 1 \\
        \midrule
        \textbf{Actor-Critic} & \\
        Imagination Horizon ($H$) & 15 \\
        Discount Horizon ($1/(1 - \gamma)$) & 333 \\
        Return Lambda ($\lambda$) & 0.95 \\
        Actor Entropy Regularizer ($\eta$) & $3 \times 10^{-4}$ \\
        \midrule
        \textbf{GRU Layer} & \\
        Hidden Units & 256 \\
        Activation Function & SiLU \\
        \bottomrule
    \end{tabular}
\end{table}

\subsection{Advantage Actor-Critic (A2C) Baseline}

To establish a robust baseline for comparison, we implemented Advantage Actor-Critic (A2C) \cite{pmlr-v48-mniha16} without modifications, using the same hyperparameters as in the original Park OS experiments \cite{mao2019park}. A2C is a widely used model-free reinforcement learning algorithm that combines policy-based and value-based methods, striking a balance between stable training and effective policy updates. A2C follows the actor-critic paradigm, where:
\begin{itemize}
    \item The actor is responsible for selecting actions by optimizing a stochastic policy $\pi(a_t | s_t; \theta)$, where $a_t$ is the selected action at state $s_t$, and $\theta$ represents the policy parameters.
    \item The critic estimates the state-value function $V(s_t; \phi)$, which provides an estimate of the long-term return from a given state, parameterized by $\phi$.
\end{itemize}

A2C improves upon traditional policy gradient methods by incorporating an advantage function to reduce variance and stabilize updates:

\begin{equation}
    A(s_t, a_t) = R_t - V(s_t; \phi)
\end{equation}

\noindent where $R_t$ is the estimated return at time $t$, and $V(s_t; \phi)$ is the value function's estimate of the expected return. This advantage function effectively normalizes updates, ensuring that actions are encouraged only when they result in better-than-expected outcomes. The A2C architecture used in this study consists of:
\begin{itemize}
    \item \textbf{Fully Connected Layers:} The input state is processed through two hidden layers of 256 neurons each, activated using ReLU.
    \item \textbf{Actor Head:} Outputs a probability distribution over available actions using a softmax layer.
    \item \textbf{Critic Head:} Outputs a scalar value estimate $V(s_t)$ for the given state using a linear layer.
\end{itemize}

The policy and value networks share feature extraction layers, following standard A2C practice. This architecture ensures efficient state encoding while enabling stable training through shared representations.

\subsection{Server Load Balancing Environment}

The server load balancing environment in Park simulates a heterogeneous multi-server system, where an RL agent must dynamically allocate incoming jobs to servers to minimize average job completion time. This task is a classic queueing problem with no known closed-form optimal solution \cite{harchol2010balance}. The heterogeneity in server processing rates and the variance in job sizes create a non-trivial scheduling challenge, requiring intelligent load-aware dispatching strategies. The environment consists of:
\begin{itemize}
    \item \textbf{Multiple servers ($k$):} The system comprises 10 servers, each with a unique processing rate that varies linearly from 0.15 to 1.05.
    \item \textbf{Poisson-distributed job arrivals:} Jobs arrive at a rate defined by a Poisson process with a configurable inter-arrival interval.
    \item \textbf{Pareto-distributed job sizes:} Job sizes follow a heavy-tailed Pareto distribution, leading to the presence of both short jobs and large jobs that heavily impact queue dynamics.
    \item \textbf{System Load:} The default configuration yields a 90\% system load, ensuring the environment remains highly dynamic and computationally constrained.
\end{itemize}

\paragraph{State Representation}
At each job arrival, the agent observes the current system state as a vector:

\begin{equation}
    s_t = (j, s_1, s_2, ..., s_k)
\end{equation}

\noindent where:
\begin{itemize}
    \item $j$ is the size of the incoming job.
    \item $s_k$ is the current queue length at server $k$.
\end{itemize}

This partial observability necessitates the use of memory-based policies, as the agent must infer longer-term workload trends from immediate observations.

\paragraph{Action Space}
The agent selects an action:

\begin{equation}
    a_t \in \{1, 2, ..., k\}
\end{equation}

\noindent where $a_t$ determines the server queue to which the incoming job is assigned. Unlike traditional scheduling heuristics, which rely on fixed rules (e.g., Join the Shortest Queue, Least Work Left), reinforcement learning enables adaptive workload-sensitive scheduling.

\paragraph{Reward Function}
The reward function encourages reducing average job completion time, given by:

\begin{equation}
    r_t = \sum_{n} [\min(t_n, c_n) - t_{n-1}]
\end{equation}

\noindent where:
\begin{itemize}
    \item $t_n$ is the time at step $n$.
    \item $c_n$ is the completion time of job $n$.
\end{itemize}

This formulation ensures that the agent is rewarded for minimizing queue delay, promoting efficient job dispatching while discouraging excessive queue buildup.

\paragraph{Challenges and Learning Opportunities}
This environment introduces several key challenges:
\begin{itemize}
    \item Balancing long and short jobs: The agent must determine when to prioritize small jobs to prevent them from being blocked by large jobs.
    \item Handling dynamic workloads: The variability in job arrival rates and sizes requires the agent to adapt to shifting traffic patterns.
    \item Optimizing under limited visibility: The agent only sees the current queue state, requiring strategic memory utilization for long-term optimization.
\end{itemize}

\subsection{Training Procedure}

To evaluate adaptability, we introduced dynamic workload variations in the Park OS environment, modifying job arrival patterns and Pareto-distributed job sizes to create three difficulty levels: Easy, Medium, and Hard (as seen in Table~\ref{tab:workload-difficulty}). These variations simulate realistic workload fluctuations, requiring agents to generalize across unseen job distributions. Each test difficulty level is characterized by three key parameters:

\begin{itemize}
    \item \textbf{Job Arrival Interval:} Defines the average time (in simulation steps) between consecutive job arrivals. Lower values indicate higher workload intensity, increasing the challenge for the scheduling agent.
    \item \textbf{Pareto Shape Parameter:} Controls the skewness of the job size distribution. A higher shape value increases the likelihood of larger jobs, making load balancing more difficult.
    \item \textbf{Pareto Scale Parameter:} Determines the absolute scale of job sizes. Higher values lead to larger overall job sizes, increasing resource contention.
\end{itemize}

\begin{table}[ht]
    \centering
    \caption{Workload Characteristics by Test Difficulty Level}
    \label{tab:workload-difficulty}
    \begin{tabular}{lccc}
        \toprule
        \textbf{Difficulty} & \textbf{Arrival Interval} & \textbf{Pareto Shape} & \textbf{Pareto Scale} \\
        \midrule
        Easy   & 100 & 1.5 & 80  \\
        Medium & 80  & 2.0 & 120 \\
        Hard   & 50  & 2.5 & 150 \\
        \bottomrule
    \end{tabular}
\end{table}

To ensure exposure to diverse workload conditions, agents were trained on 1,000 episodes, with workload parameters uniformly sampled across the defined parameter space. Training followed a three-step cycle:

\begin{enumerate}
    \item \textbf{Experience Collection:} Agents interacted with the Park environment, generating episode trajectories $(s_t, a_t, r_t, s_{t+1})$.
    \item \textbf{Policy Update:} Based on stored experiences, DreamerV3 updated its world model, policy, and value functions, while A2C updated its policy and value networks using the advantage function.
    \item \textbf{Evaluation:} After every training episode, the agent was tested across all three workload difficulty levels, recording the cumulative reward achieved in each environment.
\end{enumerate}

\begin{figure*}
  \begin{minipage}{0.49\linewidth}
    \includegraphics[width=\linewidth]{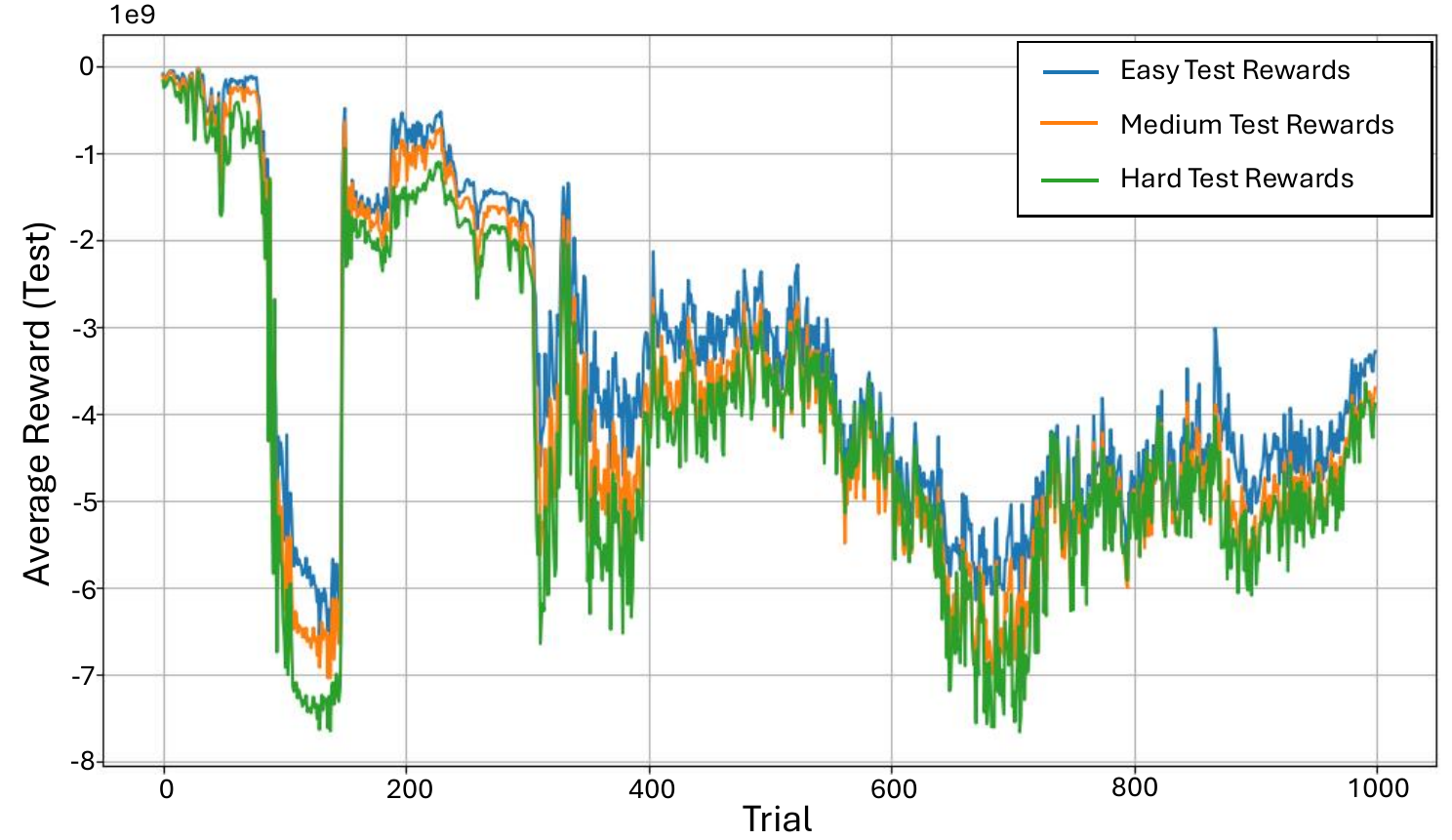}
  \end{minipage}
  \hfill
  \begin{minipage}{0.49\linewidth}
    \includegraphics[width=\linewidth]{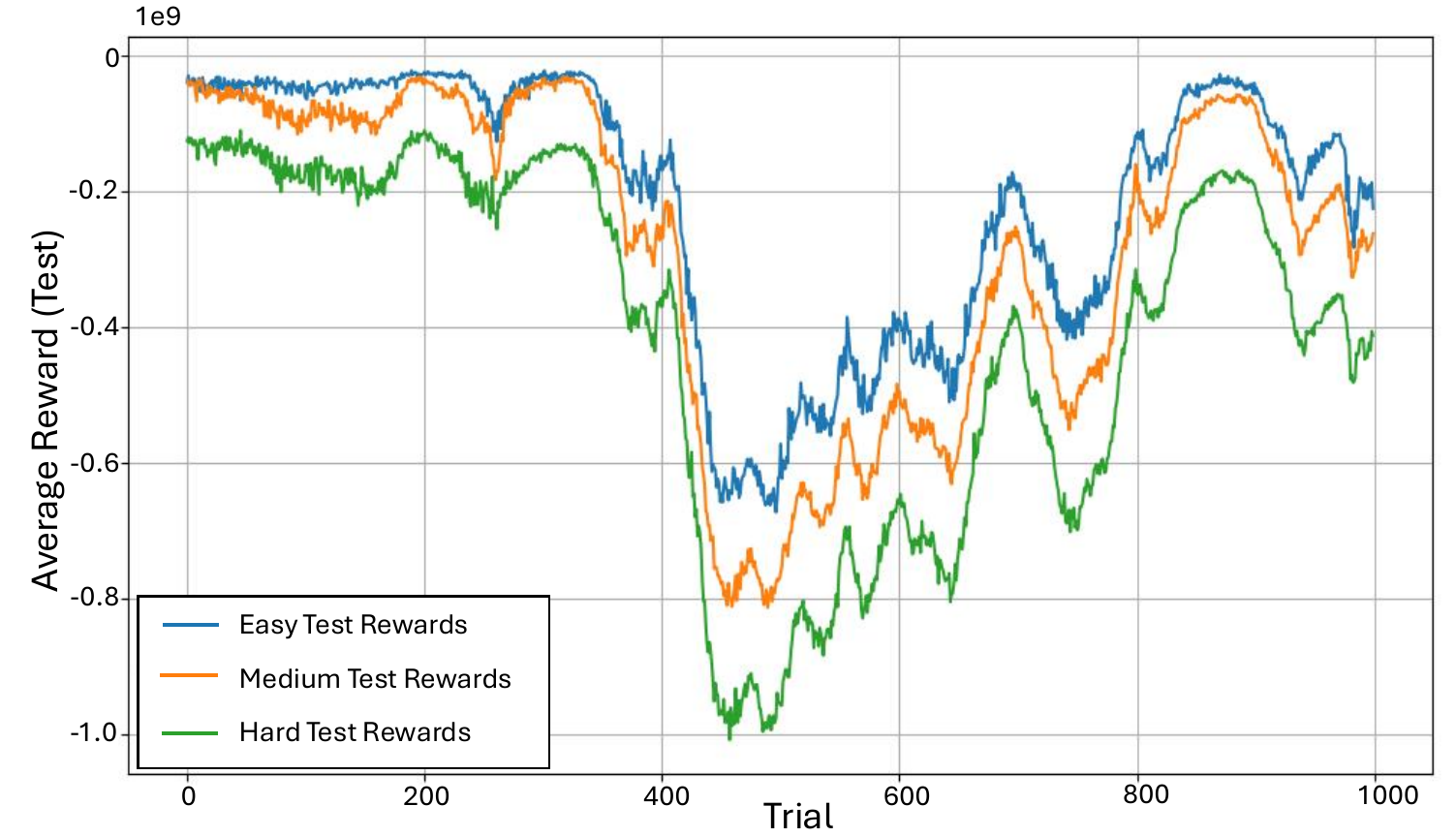}
  \end{minipage}
  \caption{Performance of A2C (Left) and DreamerV3 (Right) on Adaptive Load Balancing Tasks}
  \Description{DreamerV3 with recurrent policy network demonstrates superior resilience to catastrophic forgetting compared with the A2C algorithm.}
  \label{fig:adaptive_trials_results}
\end{figure*}

\section{Results and Analysis}
The adaptive trials reveal a stark contrast between the performance of the A2C algorithm and the DreamerV3-based RL² agent when subjected to dynamically varying workload distributions. Figure~\ref{fig:adaptive_trials_results} illustrates these differences, showing that A2C exhibited significant instability, whereas the DreamerV3-based RL² agent demonstrated strong adaptability and retention of learned behaviors.

The A2C algorithm displayed high variance across episodes, particularly when transitioning between different workload distributions. Performance degradation was evident in later training stages, with reward values steadily declining as the algorithm struggled to adjust to new workload patterns. The instability in A2C's learning process can be attributed to catastrophic forgetting, where the model fails to retain knowledge of previous workload distributions while adapting to new ones. Over time, this forgetting effect compounded, leading to a performance collapse that eventually stabilized near a reward scale of -6e-9. The sharp decline in A2C’s effectiveness underscores its inability to generalize effectively in non-stationary environments. This limitation is particularly problematic in real-world scenarios, such as operating system resource management, where workload conditions fluctuate unpredictably, requiring models that can learn continuously without erasing prior knowledge.

In contrast, the DreamerV3-based RL² agent displayed remarkable stability across the same trials. Unlike A2C, which suffered significant performance degradation, DreamerV3 maintained consistently high performance, with its reward trajectory staying well above the -1e-9 threshold. Beyond stability, the DreamerV3-based RL² agent exhibits an upward performance trajectory across episodes, indicating continuous improvement. Unlike A2C, which experienced a progressive decline, the DreamerV3 agent’s learning curve suggests steady refinement, with potential for further gains over extended training. This positive performance trend is critical for real-world dynamic environments, where workloads shift unpredictably. The ability to not only maintain but improve performance over time suggests that DreamerV3’s integration of discrete world models and recurrent learning mechanisms offers significant benefits in adaptive resource allocation tasks.

The superior performance of DreamerV3-based RL² in dynamic workload environments underscores its potential for real-world applications. Unlike A2C, which degrades under shifting workloads, DreamerV3’s approach of learning structured latent representations allows it to:

\begin{enumerate}
    \item Adapt to fluctuating job arrival rates and server loads in real-time.
    \item Minimize average job completion time while maintaining efficiency.
    \item Scale effectively without requiring extensive retraining.
\end{enumerate}

These properties make DreamerV3 highly suitable for deployment in production systems, particularly in scenarios where OS-level resource allocation, cloud computing load balancing, and real-time job scheduling require continual adaptation.

\section{Conclusion and Future Work}

This study underscores the transformative advantages of integrating discrete world models with meta-reinforcement learning algorithms like RL² for adaptive load balancing in operating systems. By leveraging DreamerV3’s discrete latent space representations, which efficiently encode complex and dynamic environmental states, and RL²’s rapid adaptation capabilities, this approach outperforms traditional reinforcement learning methods in dynamic and unpredictable workload conditions. The key benefits demonstrated include enhanced learning efficiency, robust adaptability to fluctuating workloads, and significant resilience to catastrophic forgetting. These capabilities ensure that agents can generalize effectively across diverse workload conditions while retaining and building upon prior knowledge, minimizing the need for extensive retraining in non-stationary environments.

The ability to mitigate catastrophic forgetting is particularly noteworthy, as it addresses a critical limitation in lifelong learning systems. By preserving learned behaviors and adapting seamlessly to new scenarios, this approach achieves a level of stability and robustness that is essential for real-world applications. Unlike traditional model-free methods such as A2C, which suffer from performance degradation over time, the DreamerV3-based RL² agent maintains long-term performance stability and even exhibits incremental improvements across training episodes. This suggests its potential as a scalable and forward-looking solution for managing the complexities of modern, dynamic operating systems.

Despite these promising results, several key areas remain for further investigation to enhance the applicability and robustness of this approach:

\begin{itemize}
    \item \textbf{Real-World Validation on Hardware Systems:} While the Park OS simulation provides a controlled and realistic testbed, real-world operating systems introduce additional hardware constraints, execution latencies, and unpredictable system-level interactions. Deploying this method on live operating system instances will be crucial to assessing its practical feasibility, performance trade-offs, and integration challenges.
    
    \item \textbf{Ablation Study on the Recurrent Policy Network:} The inclusion of a GRU-based recurrent policy network in the DreamerV3 architecture has been instrumental in retaining task knowledge over time. However, a rigorous ablation study is necessary to quantify its exact impact. Evaluating DreamerV3 with and without the recurrent layer will provide insights into how much temporal memory contributes to adaptation and whether alternative architectures (e.g., transformers or attention-based memory mechanisms) might yield better performance.

    \item \textbf{Scalability in Distributed and Cloud Environments:} Extending this approach to distributed operating systems and cloud computing environments will be essential for understanding its scalability in heterogeneous, multi-agent computing infrastructures. Evaluating performance in containerized workloads, multi-node scheduling, and federated resource allocation will highlight the broader applicability of meta-RL to large-scale computing systems.

    \item \textbf{Robustness to Adversarial Workloads and System Failures:} In real-world applications, workloads are not always benign—they may exhibit bursty behavior, adversarial patterns, or even system faults. Investigating how the DreamerV3-based RL² agent responds to workload anomalies, unexpected hardware failures, or adversarial job scheduling attacks will be essential for ensuring the safety, robustness, and reliability of reinforcement learning-based OS management.
    
    \item \textbf{Applications Beyond Load Balancing:} The DreamerV3-RL² framework is not limited to job scheduling and can be extended to other resource management tasks, such as memory allocation, CPU scheduling, and network bandwidth optimization. Investigating its performance across diverse OS-related decision-making processes will establish its versatility as a general-purpose reinforcement learning approach for system optimization.
\end{itemize}

By addressing these key challenges, this research can pave the way for more advanced, scalable, and resilient reinforcement learning-based solutions in operating system resource management. Future explorations into real-world deployment, architectural refinements, and broader applicability will contribute to the advancement of meta-reinforcement learning in complex and dynamic computing environments.

\balance

\begin{acks}
Thank you to Dr. Taeyeong Choi for continued encouragement and guidance. Special thanks to Github user \textit{NM512} who's PyTorch DreamerV3 implementation was a helpful reference.. 
\end{acks}

\bibliographystyle{ACM-Reference-Format}
\bibliography{Redovian-ACMSE-2025}

\end{document}